\documentclass[conference]{IEEEtran}
\IEEEoverridecommandlockouts
\usepackage{cite}
\usepackage{amsmath,amssymb,amsfonts}
\usepackage{graphicx}
\usepackage{textcomp}
\usepackage{xcolor}
\def\BibTeX{{\rm B\kern-.05em{\sc i\kern-.025em b}\kern-.08em
    T\kern-.1667em\lower.7ex\hbox{E}\kern-.125emX}}

\usepackage{algorithm}
\usepackage{algorithmicx}
\usepackage[]{algpseudocode}
\usepackage{hyperref}
\usepackage{booktabs}
\usepackage{url}

\begin{document}

\title{$b$-Bit Sketch Trie: Scalable Similarity Search on Integer Sketches}

\author{\IEEEauthorblockN{Shunsuke Kanda}
\IEEEauthorblockA{\textit{RIKEN Center for Advanced Intelligence Project}\\
Tokyo, Japan \\
shunsuke.kanda@riken.jp}
\and
\IEEEauthorblockN{Yasuo Tabei}
\IEEEauthorblockA{\textit{RIKEN Center for Advanced Intelligence Project}\\
Tokyo, Japan \\
yasuo.tabei@riken.jp}
}

\maketitle

\begin{abstract}
Recently, randomly mapping vectorial data to strings of discrete symbols (i.e., \emph{sketches}) for fast and space-efficient similarity searches has become popular. Such random mapping is called \emph{similarity-preserving hashing} and approximates a similarity metric by using the Hamming distance. Although many efficient similarity searches have been proposed, most of them are designed for binary sketches. Similarity searches on integer sketches are in their infancy. In this paper, we present a novel space-efficient trie named \emph{$b$-bit sketch trie} on integer sketches for scalable similarity searches by leveraging the idea behind \emph{succinct data structures} (i.e., space-efficient data structures while supporting various data operations in the compressed format) and a favorable property of integer sketches as fixed-length strings. Our experimental results obtained using real-world datasets show that a trie-based index is built from integer sketches and efficiently performs similarity searches on the index by pruning useless portions of the search space, which greatly improves the search time and space-efficiency of the similarity search. The experimental results show that our similarity search is at most one order of magnitude faster than state-of-the-art similarity searches. Besides, our method needs only 10 GiB of memory on a billion-scale database, while state-of-the-art similarity searches need 29 GiB of memory.
\end{abstract}


\newcommand{\Str}[1]{\texttt{#1}}
\newcommand{\Ceil}[1]{\lceil{#1}\rceil}
\newcommand{\Floor}[1]{\lfloor{#1}\rfloor}
\newcommand{\Tuple}[1]{({#1})}
\newcommand{\Order}{\mathcal{O}}

\newcommand{\StrFn}{\textsf{str}}
\newcommand{\Sigs}{\textsf{sigs}}
\newcommand{\Ham}{\textsf{ham}}
\newcommand{\Rank}{\textsf{rank}}
\newcommand{\Select}{\textsf{select}}
\newcommand{\Lexid}{\textsf{lex}}
\newcommand{\Pop}{\textsf{popcnt}}
\newcommand{\Children}{\textsf{children}}

\newcommand{\Dens}{dens}

\newcommand{\ArH}{\textsf{H}}
\newcommand{\ArC}{\textsf{C}}
\newcommand{\ArB}{\textsf{B}}

\newcommand{\ArP}{\textsf{P}}
\newcommand{\ArD}{\textsf{D}}

\newcommand{\Bin}{\textsf{bin}}

\newcommand{\dSD}{\delta}
\newcommand{\dSS}{\Delta}

\newcommand{\Trie}{\mathcal{T}}
\newcommand{\TrieSS}{\mathcal{T}_{\dSS}}

\newcommand{\DHT}{{TABLE}}
\newcommand{\LIST}{{LIST}}

\def\TrieS{{SI-$b$ST}}
\def\TrieM{{MI-$b$ST}}
\def\HashS{{SIH}}
\def\HashM{{MIH}}
\def\Hm{{HmSearch}}

\def\Louds{\textsf{Louds}}
\def\FST{\textsf{FST}}

\def\Book{{Review}}
\def\CP{{CP}}
\def\DEEP{{DEEP}}
\def\SIFT{{SIFT}}
\def\GIST{{GIST}}

\newcommand{\qref}[1]{(\ref{#1})}
\newcommand{\fref}[1]{Figure \ref{#1}}
\newcommand{\tref}[1]{Table \ref{#1}}
\newcommand{\trefs}[2]{Tables \ref{#1} and \ref{#2}}
\newcommand{\sref}[1]{Section \ref{#1}}
\newcommand{\srefs}[2]{Sections \ref{#1} and \ref{#2}}
\newcommand{\gref}[1]{Algorithm \ref{#1}}
\newcommand{\pref}[1]{Property \ref{#1}}
\newcommand{\aref}[1]{Appendix \ref{#1}}

\newcommand{\argmin}{\mathrm{arg\,min}}

\section{Introduction}

The similarity search of vectorial data in databases has been a fundamental task in recent data analysis, and it has various applications such as near duplicate detection in a collection of web pages~\cite{henzinger2006finding}, context-based retrieval in images~\cite{song2013inter}, and functional analysis of molecules~\cite{ito2012possum}.
Recently, databases in these applications have become large, and vectorial data in these databases also have been high dimensional, which makes it difficult to apply existing similarity search methods to such large databases.
There is thus a strong need to develop much more powerful methods of similarity search for efficiently analyzing databases on a large-scale.

A powerful solution to address this need is {\em similarity-preserving hashing}, which intends to approximate a similarity measure by randomly mapping vectorial data in a metric space to strings of discrete symbols (i.e., {\em sketches}) in the Hamming space.
Early methods include Sim-Hash for cosine similarity~\cite{Gionis1999similarity}, which intends to build binary sketches from vectorial data for approximating cosine similarity.
Quite a few similarity searches for binary sketches have been proposed thus far~\cite{gog2016fast,liu2011large,manku2007detecting,norouzi2012fast,norouzi2014fast,qin2018gph,torralba2008small}.
Recently, many types of similarity-preserving hashing algorithms intending to build sketches of non-negative integers (i.e., {\em $b$-bit sketches}) have been proposed for efficiently approximating various similarity measures. 
Examples are $b$-bit minwise hashing (minhash)~\cite{broder1997syntactic, theobald2008spotsigs,li2010b} for Jaccard similarity, 0-bit consistent weighted sampling (CWS) for min-max kernel~\cite{li20150}, and 0-bit CWS for generalized min-max kernel~\cite{li2017linearized}.
Thus, developing scalable similarity search methods for $b$-bit sketches is a key issue in large-scale applications of similarity search. 

Similarity searches on binary sketches are classified into two approaches: \emph{single-} and \emph{multi-indexes}.
Single-index (e.g., \cite{torralba2008small,norouzi2014fast}) is a simple approach for similarity searches and builds an inverted index whose key is a sketch in a database and value is the identifiers with the same sketch.
The similarity search for a query sketch is performed by generating all the sketches similar to the query as candidate solutions and then finding the solution set of sketches equal to a generated sketch by retrieving the inverted index. 
Typically, the hash table data structure is used for implementing the inverted index.
A crucial drawback of single-index is that the query time becomes large for sketches with a large alphabet and a large Hamming distance threshold because the number of generated sketches is exponentially proportional to the alphabet size of sketches and the Hamming distance threshold.

To overcome the drawback of the single-index approach, the multi-index approach \cite{greene1994multi} has been proposed as a generalization of single-index for faster similarity searches, and it has been studied well for the past few decades~\cite{gog2016fast,liu2011large,manku2007detecting,norouzi2012fast,qin2018gph,qin2018pigeonring}. 
Multi-index divides input sketches into several blocks of short sketches of possibly different lengths and builds inverted indexes from the short sketches in each block, where the key of an inverted index is a short sketch in each block, and its value is the identifier of the sketch with the short sketch. 
As in single-index, the inverted indexes are implemented using a hash table data structure. 
The similarity search of a query consists of two steps: filtering and verification. 
The filtering step divides the query into several short sketches in the same manner and finds short sketches similar to a short sketch of the query in each block by retrieving the inverted index. 
After that, the verification step removes false positives (i.e., a pair of short sketches in a block is similar but the corresponding pair of their original sketches is dissimilar) from those candidates by computing the Hamming distance between the pair of every candidate and query in the verification step. 
Although the candidate solutions for short sketches in each block are generated in multi-index and are retrieved by the inverted index as in single-index, the number of candidate solutions generated in each block becomes smaller, resulting in faster similarity searches especially when a large threshold is used. 

Many methods using the multi-index approach have been proposed for scalable similarity searches for binary sketches.
Although some recent methods \cite{qin2018pigeonring,qin2018gph,gog2016fast} have successfully improved the verification step in the multi-index approach, they have a serious issue when applied to $b$-bit sketches because the computation time of the filtering step is exponentially proportional to the alphabet size (i.e., the value of $b$) in $b$-bit sketches. 
Although Zhang et al. \cite{zhang2013hmsearch} have tried to improve the filtering step for $b$-bit sketches, their method has a scalability issue. 
Since many similarity preserving hashing algorithms for $b$-bit sketches have been proposed for approximating various similarity measures, 
an important open challenge is to develop a scalable similarity search for $b$-bit sketches. 

\emph{Trie} \cite{fredkin1960trie} is an ordered labeled tree data structure for a set of strings and supports various string operations such as string search and prefix search with a wide variety of applications in string processing such as string dictionaries \cite{kanda2017compressed}, $n$-gram language models \cite{pibiri2017efficient}, and range query filtering \cite{zhang2018surf}.
A typical pointer-based representation of trie consumes a large amount of memory. 
Thus, recent researches have focused on space-efficient representations~\cite{jacobson1989space,delpratt2006engineering,zhang2018surf}.
To date, trie has been applied only to the limited application domains listed above.
However, as we will see, there remains great potential for a wide variety of applications.

\emph{Contribution:}
In this paper, we present a novel trie representation for $b$-bit sketches, which we call \emph{$b$-bit Sketch Trie ($b$ST)}, to enhance the search performance of single-index and multi-index. 
We design $b$ST by leveraging a \emph{succinct data structure}~\cite{jacobson1989space} (i.e, a compressed data structure while supporting fast data operations in the compressed format) and a favorable property behind $b$-bit sketches as fixed-length strings.
Our similarity search method represents a database of $b$-bit sketches in $b$ST and solves the Hamming distance problem for a given query by traversing the trie while pruning unnecessary portions of the search space.
We experimentally test our similarity search using $b$ST's ability to retrieve massive databases of $b$-bit sketches similar to a query and show that our similarity search performs superiorly with respect to scalability, search performance, and space-efficiency.

\section{Similarity Search Problem}

We formulate the similarity search problem for $b$-bit sketches. 
A $b$-bit sketch is an $L$-dimensional vector of integers, each of which is within range $[1,2^b]$, and it is also equivalent to a string of length $L$ over alphabet $\Sigma=[1,2^b]$. 
We also denote elements in $\Sigma$ (i.e., characters) by the small English letters (e.g., \Str{a}, \Str{b}, and \Str{c}) in the examples of this paper. 
A database of $b$-bit sketches consists of $n$ $b$-bit sketches $s_1,s_2,\ldots,s_n$, where $s_i\in \Sigma^L$. 
Given $b$-bit sketch $q$ and threshold $\tau$ as a query, 
the task of the similarity search is to report all the identifiers ${I}$ of $b$-bit sketches $s_1,s_2,\ldots,s_n$ whose Hamming distance to $q$ is 
no more than $\tau$, i.e., ${I}=\{i : \Ham(s_i, q) \leq \tau\}$, where $\Ham(\cdot, \cdot)$ denotes the Hamming distance
between two strings (i.e., the number of positions at which the corresponding characters between two strings are different).

\section{Related Works}
\label{sect:relate}

Many methods for similarity searches on binary sketches have been proposed, and they are classified into two approaches: single- and multi-indexes.
Theoretical aspects of similarity search have also been argued  \cite{belazzougui2009faster,belazzougui2012compressed,cole2004dictionary,chan2010compressed,yao1997dictionary}.
The following subsections review practical similarity search methods. 

\subsection{Single-Index Approach}

Single-index (e.g., \cite{torralba2008small,norouzi2014fast}) is a simple approach for the similarity search.
This approach typically builds an inverted index implemented using a hash table data structure.
From a database of sketches $s_1,s_2,\ldots,s_n$, it builds an inverted index whose key is sketch $s_i$ in the database and value is the set of identifiers with the same sketch. 
The similarity search for given query $q$ and threshold $\tau$ is performed by generating the set $Q$ of all the sketches $q^\prime$ similar to query $q$ (i.e., $Q=\{q^\prime \in \Sigma^L : \Ham(q,q^\prime) \leq \tau \}$) and then finding the solution set $I$ of sketches equal to generated sketch $q^\prime \in Q$ (i.e., $I=\{i : \exists q^\prime \in Q 
~{\mbox s.t.}~ s_i = q^\prime \}$) by retrieving the inverted index. 
Each element in set $Q$ is called \emph{signature}, and 
single-index using the hash table data structure is referred to as {\em single-index hashing (SIH)}.

The search time of SIH is evaluated by using the retrieval time for the signatures in $Q$ and access time for solution set $I$ (see \aref{app:sih} for detailed analysis), and it is linearly proportional to $L$ and exponentially proportional to $\tau$ and $b$.   
Since the number of signatures can exceed that of sketches in the database for large parameters $b$, $L$, and $\tau$, 
a naive linear search can be faster than SIH. 
In particular, for $b > 1$ (i.e., non-binary sketches), the time performance is much more sensitive to $\tau$, resulting in difficulty in applying SIH to $b$-bit sketches when a large $\tau$ is used.

\subsection{Multi-Index Approach}
\label{sect:multi}

The multi-index approach partitions sketch $s_i$ for each $i = 1,2,\ldots,n$ in a database into $m$ blocks (i.e., substrings) $s_i^1,s_i^2,\ldots,s_i^m$ of lengths $L^1,L^2,\ldots,L^m$, respectively.
The $m$ blocks are disjoint.
The approach builds inverted index $X^j$ using the $j$-th block $s_1^j,s_2^j,\ldots,s_n^j$ for each $j = 1,2,\ldots,m$, where the key of $X^j$ is the $j$-th block $s_i^j$ for each $i= 1,2,\ldots,n$, and the value of $X^j$ is the set of identifiers of the original sketch with the same block $s_i^j$.

A query is searched in two steps: filter and verification.
Given query sketch $q$ and threshold $\tau$, $q$ is partitioned into $m$ blocks  $q^1,q^2,\ldots,q^m$ of lengths $L^1,L^2,\ldots,L^m$, respectively.
Thresholds $\tau^1,\tau^2,\ldots,\tau^m$ no more than $\tau$ are assigned to $m$ blocks. 
Note that $\tau^j = \Floor{\tau/m}$ for $j=1,2,\ldots,m$ is traditionally used to avoid false negatives on the pigeonhole principle (e.g., \cite{zhang2013hmsearch,gog2016fast}).
At the filter step, the set $C^j$ of candidate solutions for $q^j$ for each $j = 1,2,\ldots,m$ is obtained by retrieving inverted index $X^j$. 
As in the single-index approach, the set $Q^j$ of all the sketches similar to $q^j$ (i.e. $Q^j=\{q^\prime\in \Sigma^{L^j}: \Ham(q^j,q^\prime)\leq \tau^j\}$) for each $j= 1,2,\ldots,m$ is generated, and  
$C^j$ is computed by retrieving inverted index $X^j$ for each $q^\prime \in Q^j$. 
The verification step removes false positives  (i.e., $\{i: \Ham(s_i, q)\geq \tau, i\in C^j\}$) by computing the Hamming distance 
between the pair of sketches $s_i$ for each $i\in C^j$ and query $q$. 
See \aref{app:sih} for detailed analysis of the search performance of the multi-index approach.

Quite a few similarity searches based on the multi-index approach have been proposed.
We briefly review some state-of-the-art methods as follows.
Gog and Venturini \cite{gog2016fast} proposed an efficient multi-index implementation method, which exploits succinct data structures \cite{jacobson1989space} and triangle inequality.
Qin et al. \cite{qin2018gph} generalized the pigeonhole principle and proposed a method that assigns variable thresholds $\tau^j$ for each $j = 1,2,\ldots,m$ by considering the distribution of sketches.
Qin and Xiao \cite{qin2018pigeonring} proposed the {pigeonring principle} to shorten the verification time by exploiting the sum of thresholds assigned to adjacent blocks and constraining the number of candidate solutions (i.e., the size of $C^j$).
The efficiency of the pigeonring principle is verified for long sketches (e.g., $L \geq 256$), enabling the assignment of a sufficient number of blocks \cite{qin2018pigeonring}.

\emph{Multi-index hashing (MIH)} \cite{norouzi2014fast} is a state-of-the-art multi-index approach using the hash table data structure for implementing an inverted index.
MIH partitions sketches $s_i$ into $m$ blocks of equal length (i.e., $L^j = \Floor{L/m}$) and assigns threshold $\tau^j = \Floor{\tau/m} - 1$ to the first $\tau - m\Floor{\tau/m}+1$ blocks and $\tau^j = \Floor{\tau/m}$ to the other blocks.

\emph{HmSearch} \cite{zhang2013hmsearch} is a representative multi-index approach originally designed for $b$-bit sketches.
HmSearch partitions sketches into blocks by using the pigeonhole principle where the length of blocks is determined such that threshold $\tau^j$ for each block is zero or one.
To avoid generating many signatures at the filter step, HmSearch  registers 
all the signatures generated from each sketch in a database to the inverted index, resulting in a large memory consumption. 
Although many similarity search methods applicable to $b$-bit sketches have been proposed (\cite{manku2007detecting,li2008efficient,liu2011large}), Zhang et al. \cite{zhang2013hmsearch} show that HmSearch performs best.

Despite the importance of similarity searches for $b$-bit sketches, no previous work has been able to achieve both fast similarity search and space-efficiency. 
The problem behind the existing methods consists of (i) inefficient similarity searches because of a large number of generated signatures, 
(ii) a large verification cost in the multi-index approach, and (iii) a large space consumption for storing multiple inverted indexes. 
We solve this problem by presenting a trie-based similarity search method and a space-efficient trie representation, named $b$ST, tailored for $b$-bit sketches.
Our method with $b$ST can be used instead of the inverted index using the hash table data structure in the single- and multi-index approaches.
Since our method solves the similarity search problem for $b$-bit sketches by traversing the trie without signature generation, fast similarity searches can be performed even for a large threshold $\tau$.
Since $b$ST compactly stores a massive database of $b$-bit sketches, 
$b$ST enables scalable similarity searches. 
The details of the proposed method are presented in the next sections.
\section{Trie-based Similarity Search}

A key idea behind our similarity search is to build $b$ST so that it supports data operations in the inverted index and to solve the similarity search problem on $b$-bit sketches by traversing $b$ST for computing Hamming distances.
$b$ST can be used instead of the inverted index in the single-index approach and in the filtering step in the multi-index approach, which result in a fast and space-efficient similarity search for $b$-bit sketches.
In this section, we first introduce a data structure of trie and then present a similarity search on a trie using pointers (PT).

\subsection{Data Structure}

\begin{figure}[tb]
    \centering
    \includegraphics[scale=0.4]{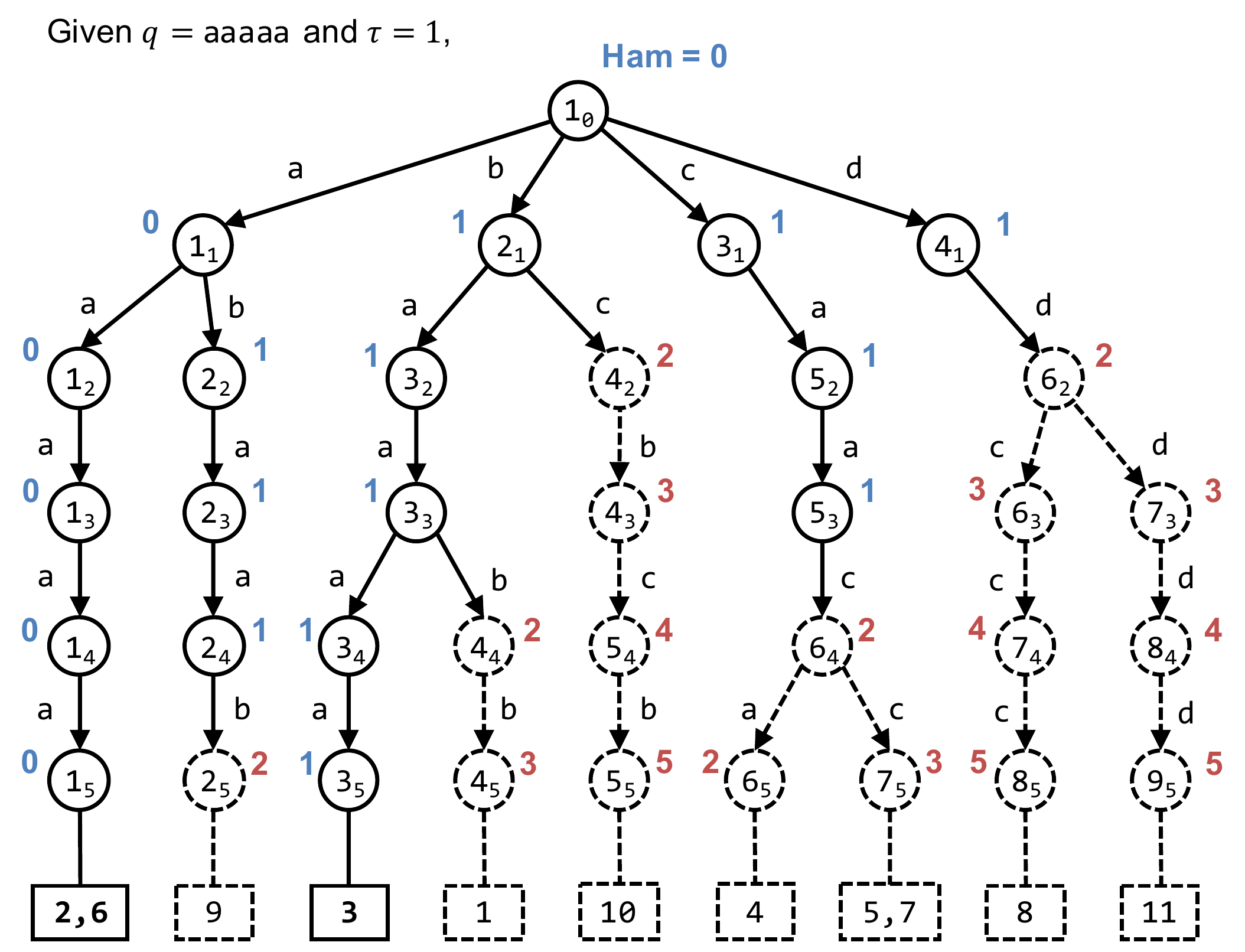}
    \caption{
        Illustration of a trie built from eleven 2-bit sketches of \Str{baabb}, \Str{aaaaa}, \Str{baaaa}, \Str{caaca}, \Str{caacc}, \Str{aaaaa}, \Str{caacc}, \Str{ddccc}, \Str{abaab}, \Str{bcbcb}, and \Str{ddddd}.
        $L = 5$.
        Each leaf has the sketch ids associated with sketches. 
        The Hamming distance between query $q=\Str{aaaaa}$ and the prefix of each node is represented by red/blue numbers.
        Solid circles (dashed circles) represent traversed nodes (pruned nodes) for $q$ and $\tau=1$.
    }
    \label{fig:trie}
\end{figure}

Trie is an ordered labeled tree representing a set of sketches (see \fref{fig:trie}).
Each node is associated with the common prefix of a subset of input sketches, and the root node (i.e., the node without a parent) is not associated with any prefix. 
Each leaf (i.e., each node without any children) is associated with input sketches of the same characters and contains their identifiers.
Each edge has a character of sketches as a label.
All outgoing edges of a node are labeled with distinct characters.

The set of prefixes at each level $\ell$ consists of substrings of length $\ell$ and is assumed to be lexicographically sorted in ascending order. 
Each node (represented by solid/dashed circles in \fref{fig:trie}) is associated with the unique node identifier (id) that is represented as notation $u_\ell$ 
by level $\ell$ and lexicographic order $u$ of the prefix from left to right at level $\ell$. 
Note that level $\ell$ is in $[0,L]$, i.e., the root id is $1_0$.
The $u$-th prefix at level $\ell$ is denoted by $\StrFn(u_\ell)$.

For the trie in \fref{fig:trie}, node $3_3$ represents the lexicographic order of $\StrFn(3_3) = \Str{baa}$, which is the third node by following $\StrFn(1_3) = \Str{aaa}$ and $\StrFn(2_3) = \Str{aba}$.

\subsection{Similarity Search}
\label{sect:trie:sim}
The similarity search for query sketch $q$ and threshold $\tau$ on a trie represented by PT is performed by traversing the trie from root $1_0$ to every leaf $u_L$ in a depth-first manner while computing the Hamming distance between $q$ and the prefix at each node.
Algorithm~\ref{alg:simsearch} shows the pseudo-code of the similarity search on PT. 
First, we initialize solution set $I$ to an empty set.
We start at root $u_\ell = 1_0$ on PT and distance count $dist=0$.
The similarity search is recursively performed by the following three steps: 
(i) given node $u_\ell$, we compute $\Children(u_\ell)$ that is a function returning set $K$ of all the pairs $(v_{\ell+1}, c)$ of child $v_{\ell+1}$ and edge label $c$ connecting $u_\ell$ and $v_{\ell+1}$,
(ii) we recursively go down to each child $v_{\ell+1}$ in $K$ if $\Ham(\StrFn(v_{\ell+1}),q[1..(\ell+1)])$ is no more than $\tau$, where $q[i..j]$ denotes the substring $q[i] q[i+1] \ldots q[j]$ for $1 \leq i \leq j \leq L$, and
(iii) if $v_{\ell+1}$ is a leaf, we add the ids associated with $v_{\ell+1}$ to solution set $I$. 
In step~(ii), we stop going down to all the descendants under node $v_{\ell+1}$ if $\Ham(\StrFn(v_{\ell+1}),q[1..(\ell+1)])$ is more than $\tau$ 
without missing all the solutions.

\begin{algorithm}[tb]
\caption{Similarity search. $q$: query sketch, $\tau$: Hamming distance threshold, $u_\ell$: $u$-th node at level $\ell$ in PT, $dist$: the Hamming distance at $u_\ell$ for $q$, and $I$: solution set of ids.}
\label{alg:simsearch}
\begin{algorithmic}[1]
\State Initialize $I \gets \emptyset$
\State{\textsc{SimSearch}{($1_{0}$, $0$)}}
\Procedure{SimSearch}{$u_\ell$, $dist$}
\If{$dist > \tau$}  \Comment{Hamming distance is more than $\tau$}
\State \Return
\EndIf
\If{$\ell = L$} \Comment{Reach leaf node}
\State{Add the ids associated with $u_\ell$ to $I$} 
\State \Return
\EndIf
\State Compute the set $K$ of pairs $(v_{\ell+1}, c)$ by $\Children(u_\ell)$
\For{each pair $(v_{\ell+1}, c)$ in $K$}
\If{$c \neq q[\ell+1]$}　
\State \textsc{SimSearch}{($v_{\ell+1}$, $dist + 1$)} 
\Else
\State{\textsc{SimSearch}{($v_{\ell+1}$, $dist$)}}
\EndIf
\EndFor
\EndProcedure
\end{algorithmic}
\end{algorithm}

Let $t^{tra}$ be the number of traversed nodes.  
The time complexity is $\Order({t}^{tra} + |I|)$, where $|I|$ denotes the cardinality of set $I$.
Let $t$ be the number of nodes in a trie.
${t}^{tra}$ can be much smaller than $t$ when a small threshold $\tau$ is used.
However, the space complexity is $\Order(t\log{t}+t \cdot b)$ bits, and it is not feasible for PT to represent tries built from massive databases.

A crucial observation of the similarity search in Algorithm~\ref{alg:simsearch} is that trie should support the $\Children$ operation for the implementation. 
Thus, we present $b$ST supporting $\Children$ for the scalable similarity search in the next section. 

\subsection{Review on Succinct Tries}

A solution for compactly representing tries is to leverage succinct trie representations. 
We review two representative succinct tries in this subsection. 

\emph{Level-order unary degree sequence trie (LOUDS-trie)} \cite{jacobson1989space,delpratt2006engineering} is a succinct representation for tries and has a wide variety of applications (e.g., \cite{kudo2011efficient,tabei2012succinct}).
LOUDS-trie represents a trie using $(b+2) \cdot t + o(t)$ bits of space, which is much smaller than $\Order(t\log{t} + t\cdot b)$ bits of space for PT, 
while supporting trie operations (e.g., computations of a child or the parent for a node) in constant time. 
\emph{Fast succinct trie (FST)}~\cite{zhang2018surf} is a practical 
variant of LOUDS-trie. 
FST consists of two LOUDS structures: one is fast and the other is space-efficient.
FST divides a trie into two layers at a certain level and builds the fast structure at the top layer and the space-efficient structure at the bottom layer.
Although the space and time complexities of FST are the same as those of LOUDS-trie, FST is smaller and much faster in practice due to the layer-wise representation of trie.

LOUDS-trie and FST are not effective to manage $b$-bit sketches because they are designed for general strings and do not consider the favorable properties of $b$-bit sketches (e.g., strings of fixed-length $L$ and the fixed-size $2^b$ of each character).
In the next section, we present a novel trie representation $b$ST, which is designed for storing $b$-bit sketches space-efficiently.

\section{$b$-Bit Sketch Trie ($b$ST)}
\label{sect:sketch-trie}
In this section, we present $b$ST, a space-efficient representation of trie for large sets of $b$-bit sketches, that supports the $\Children$ function for scalable similarity searches. 
The $b$ST design leverages an idea behind data distribution of $b$-bit sketches and 
{\em succinct data structures} \cite{jacobson1989space}, which are compressed data structures while supporting various data operations in the compressed format.

$b$-bit sketches are random strings such that the character at each position is equally distributed.
This is a major difference from strings in the natural language where a character in each position appears with a bias. 
Thus, a trie built from $b$-bit sketches has the property that the higher the level the trie nodes are at, the more children they have. 
We design a compact trie representation as $b$ST by leveraging this property. 

\begin{figure}[tb]
    \centering
    \includegraphics[scale=0.35]{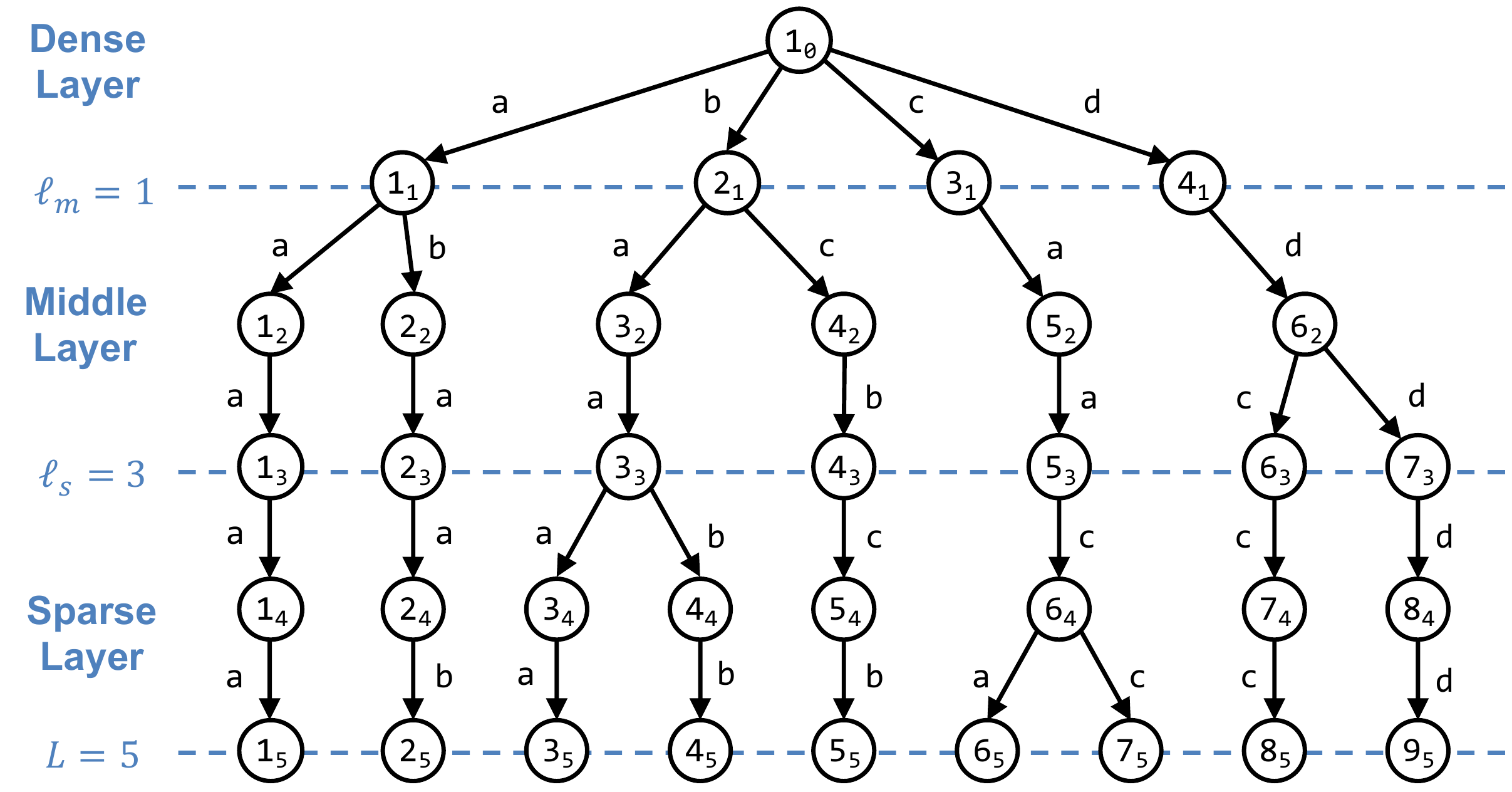}
    \caption{Illustration of the division of a trie topology into three layers of dense, middle, and sparse layers.}
    \label{fig:bST}
\end{figure}

The key ideas behind $b$ST are (i) to divide a trie topology into three layers including subtries from the top level to the bottom level according to node density, where the node density in a layer is defined as the proportion of the number of nodes at the top level to the number of nodes at the bottom level, and (ii) to apply an optimal encoding into each set of subtries in each layer (see \fref{fig:bST}).

Formally, $\ell_1$ ($\ell_2$) is the top level (the bottom level) in a layer, 
and $\ell_1 < \ell_2$. 
The node density $D(\ell_1,\ell_2)$ in the layer from  $\ell_1$ to $\ell_2$ is defined as 
\begin{equation}\label{eq:dense}
    D(\ell_1,\ell_2) = \frac{t_{\ell_2}}{t_{\ell_1}}, 
\end{equation}
where $t_{\ell_1}$ ($t_{\ell_2}$) is the number of nodes in the top level $\ell_1$ (the number of nodes in the bottom level $\ell_2$). 
The layers consist of 
(i) {\em dense layer}, (ii) {\em sparse layer}, and (iii) {\em middle layer} 
and are determined according to node densities as follows: 
(i) the dense layer is a layer from the top level (i.e., level $\ell=0$) to the maximum level $\ell_{m}$ satisfying density condition $D(0,{\ell_{m}}) = 2^{b\ell_m}$ for given $b$,
(ii) the sparse layer is a layer from the bottom level $L$ (i.e., all the nodes are leaves at level $L$) to the minimum level $\ell_{s}$ ($\geq \ell_m$) satisfying density condition $D(\ell_s, L) < \lambda$ for parameter $ \lambda \in (0,1)$, and
(iii) the middle layer is the remaining layer except for the dense and sparse layers (i.e., the layer between level $\ell_{m}$ to level $\ell_{s}$).

We present novel compact representations for subtries in dense, sparse, and middle layers in the following subsections. 

\emph{Rank and Select Data Structures:}
$b$ST leverages \Rank{} and \Select{} data structures \cite{jacobson1989space} that 
are succinct data structures on a bit array and supports
rank and selection operations on bit array $B$ of length $N$ as follows:
\begin{itemize}
    \item {$\Rank(B,i)$} returns the number of occurrences of bit \Str{1} in $B[1..i]$.
    \item {$\Select(B,i)$} returns the position in $B$ of the $i$-th occurrence of bit \Str{1}; however, if $i$ exceeds the number of \Str{1}s in $B$, it always returns $N+1$.
\end{itemize}
Given a bit array $B = [\Str{01101011}]$, $\Rank(B,5) = 3$ and $\Select(B, 4) = 7$.

The operations can be performed in $\Order(1)$ time by using auxiliary data structures of only $o(N)$ additional bits \cite{jacobson1989space}.
In our experiments, we implemented \Rank{} and \Select{} using the \emph{succinct data structure library} \cite{gog2014theory}.

\subsection{Representation for Dense Layer}
\label{sect:sketch-trie:pd}

The dense layer between level $0$ and level $\ell_{m}$ includes a complete $2^b$-ary trie of height $\ell_m$ and with $2^{b \ell_m}$ leaves. 
A characteristic property of the complete $2^b$-ary trie is that we can compactly represent it by storing only level information $\ell_m$ rather than 
complete information on the trie such as topology and edge labels.
Thus, the space usage for storing a complete $2^b$-ary trie with height $\ell_m$ is $\Order(\log{\ell_m})$. 

Given node $u_\ell$ such that $\ell<\ell_m$, $\Children(u_\ell)$ computes $2^b$ pairs $(v_{\ell+1}, c)$ of children $v_{\ell+1}$ and edge labels $c$. 
The operation returns $\{((v+1)_{\ell+1}, 1), ((v+2)_{\ell+1}, 2), \ldots, ((v+2^b)_{\ell+1}, 2^b) \}$\ where $v = (u-1) \cdot 2^{b}$.

For the trie in \fref{fig:bST}, $\Children(1_0)$ returns \{$(1_1, \Str{a})$, $(2_1, \Str{b})$, $(3_1,\Str{c})$, $(4_1,\Str{d})$\}. 
This is because $\Children(1_0)$ is computed as \{$(1_1, 1)$, $(2_1, 2)$, $(3_1, 3)$, $(4_1,4)$\}, and $1$, $2$, $3$, and $4$ correspond to $\Str{a}$, $\Str{b}$, $\Str{c}$, and $\Str{d}$, respectively. 

\subsection{Representation for Middle Layer}
\label{sect:sketch-trie:md}
The middle layer includes dense and sparse nodes. 
We present two representations of \emph{\DHT{}} and \emph{\LIST{}} for the nodes at the middle layer.
Either {\DHT{}} or {\LIST{}} is adaptively selected according to the node density at each level and is applied to the nodes.

\emph{\DHT{} Representation:}
We use bit array $\ArH_{\ell}$ of length $2^b \cdot t_{\ell-1}$ for compactly representing the nodes at level $\ell\in [\ell_m+1,\ell_s]$. 
The $((u-1)\cdot 2^b + c)$-th position on $\ArH_{\ell}$ (i.e., $\ArH_{\ell}[(u-1)\cdot 2^b + c]$) is $\Str{1}$ if and only if each node $u_{\ell-1}$ at level $\ell-1$ has a child with edge label $c$. 
Thus, each position on $\ArH_{\ell}$ represents whether there exists edge label $c$ connecting 
from node $u_{\ell-1}$ at level $\ell-1$ to a child at level $\ell$. 

\begin{figure}[tb]
    \centering
    \includegraphics[scale=0.4]{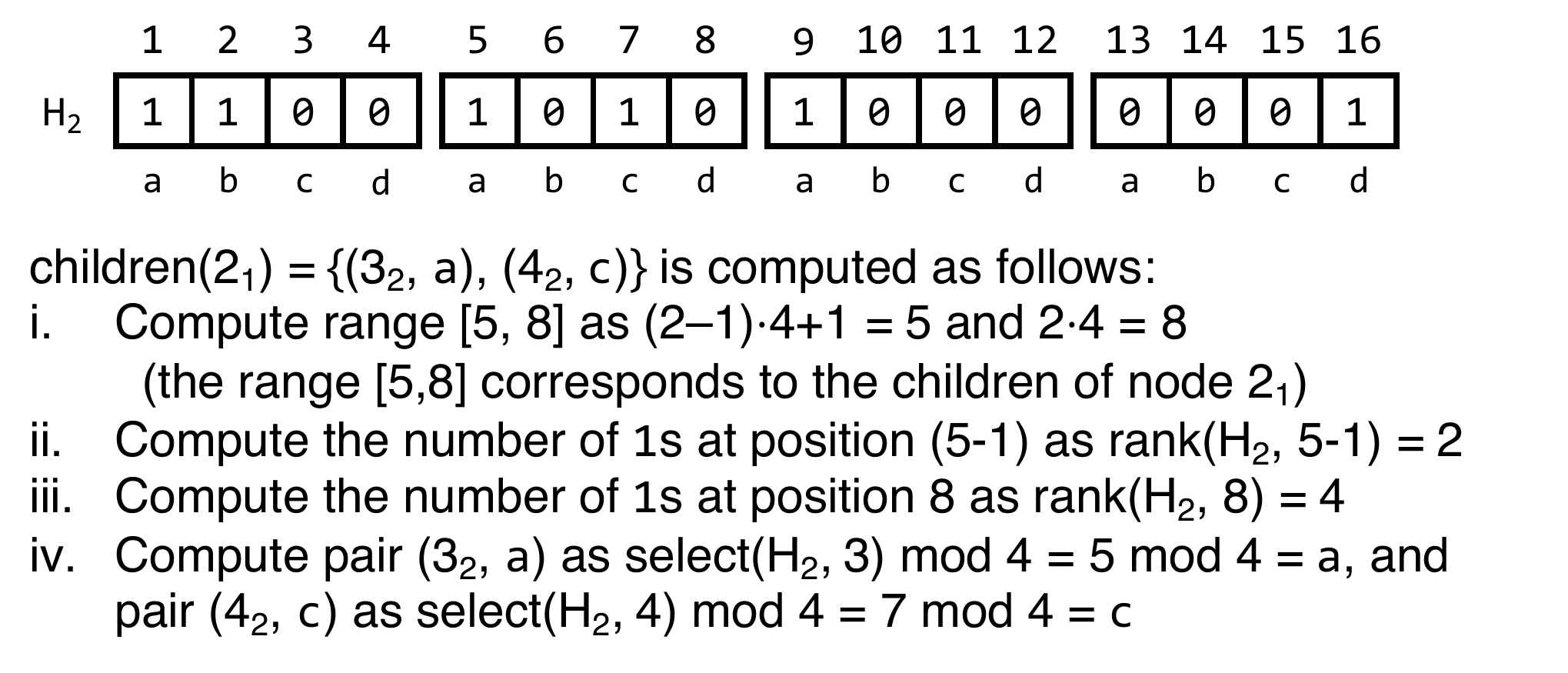}
    \caption{Illustration of \DHT{} representation for nodes at level 2 of the trie in \fref{fig:bST} and computation example of $\Children(2_1)$.}
    \label{fig:table}
\end{figure}

Array $\ArH_2$ in \fref{fig:table} shows the \DHT{} representation of nodes at level $2$ of the trie 
in \fref{fig:bST}. 
Node $2_1$ in the trie has edge labels of \Str{a} and \Str{c} connecting $2_1$'s children of $3_2$ and $4_2$, respectively. 
Thus, the first and second positions in $\ArH_2$ are $5$ and $7$, respectively. 

Given node $u_{\ell-1}$, $\Children(u_{\ell-1})(=K)$ is computed as follows:
(i) compute range $[i,j]$ on $\ArH_{\ell}$ as $i \gets (u - 1) \cdot 2^b + 1$ and $j \gets u \cdot 2^b$,
(ii) compute the number of $\Str{1}$s at position $(i-1)$ as $x \gets \Rank(\ArH_{\ell},i-1)$ if $i > 1$ or $x \gets 0$ otherwise, 
(iii) compute the number of $\Str{1}$s at position $j$ as $y \gets \Rank(\ArH_{\ell}, j)$, and 
(iv) compute the set $K$ of pairs of child id $v_\ell$ and edge label $c$ as $c \gets \Select(\ArH_{\ell}, v) \bmod 2^b$ for all $v\in [x+1,y]$.

The algorithm is made possible because range $[i,j]$  corresponds to the children of $u_{\ell-1}$, and there is a 1-to-1 correspondence between the \Str{1}s on $\ArH_{\ell}[i..j]$ and the children of $u_{\ell-1}$.

The bottom of \fref{fig:table} shows an example for computing $\Children(2_1)$.
In this example, $2^b = 4$.
Given node $2_1$, range $[i,j]$ is computed as $i \gets (2-1)\cdot 4 + 1 = 5$ and $j \gets 2 \cdot 4 = 8$.
The number of \Str{1}s at the fourth position is computed as $x \gets \Rank(\ArH_{2},5-1) = 2$.
The number of \Str{1}s at the eighth position is computed as $y \gets \Rank(\ArH_{2}, 8) = 4$.
Pair $(3_2,\Str{a})$ in $K$ is computed as $x + 1 = 3$ and $\Select(\ArH_2, 3) \bmod 4 = 5 \bmod 4 = \Str{a}$, and pair $(4_2,\Str{c})$ in $K$ is computed as $y = 4$ and $\Select(\ArH_2, 4) \bmod 4 = 7 \bmod 4 = \Str{c}$. Thus, $K=\{(3_2,\Str{a}), (4_2, \Str{c})\}$.

The space usage of \DHT{} for nodes at level $\ell$ is $2^b \cdot t_{\ell-1} + o(2^b \cdot t_{\ell-1})$ bits.

\begin{figure}[tb]
    \centering
    \includegraphics[scale=0.4]{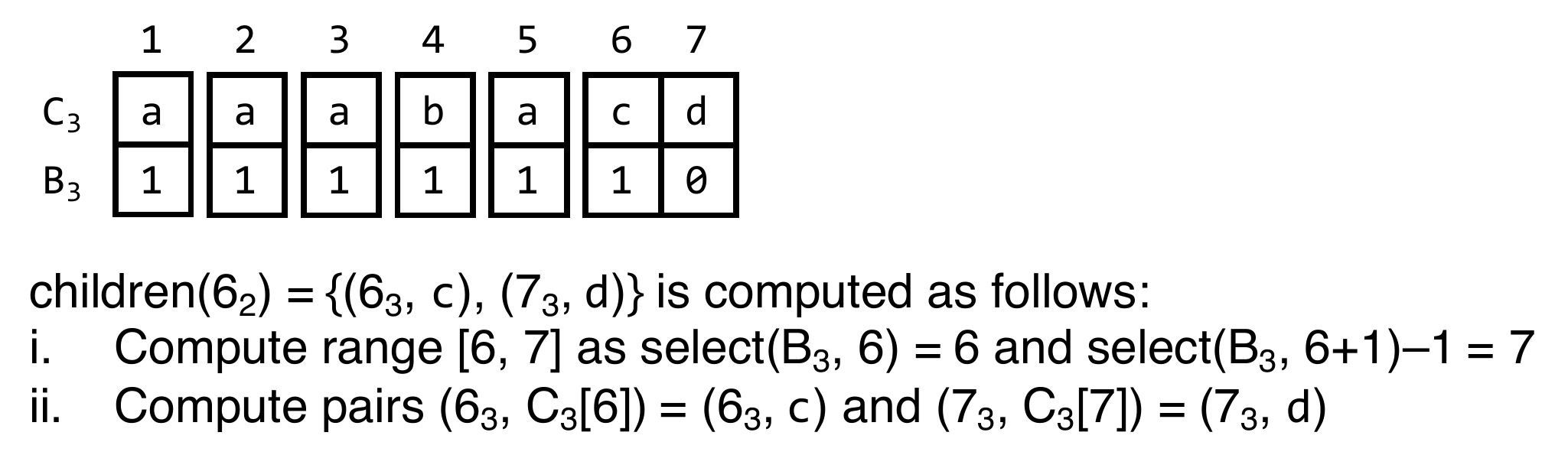}
    \caption{Illustration of \LIST{} representation for nodes at level 3 of the trie in \fref{fig:bST} and computation example of $\Children(6_2)$.}
    \label{fig:list}
\end{figure}

\emph{\LIST{} Representation:}
\LIST{} represents nodes at level $\ell \in [\ell_m+1, \ell_s]$ using array $\ArC_\ell$ of length $t_\ell$ and bit array $\ArB_\ell$ of length $t_\ell$. 
For each node $u_\ell$ at level $\ell$, $\ArC_\ell[u]$ stores the edge label between $u_\ell$ and its parent.
$\ArB_\ell[u]$ stores bit \Str{1} (i.e., $\ArB_\ell[u]=\Str{1}$) if $u_\ell$ is the first of its siblings 
(i.e., $u$ is the smallest id for its siblings) or bit $\Str{0}$ otherwise.
Arrays $\ArC_3$ and $\ArB_3$ in \fref{fig:list} are the LIST representation at level $3$ of the trie in \fref{fig:bST}.

Given node $u_{\ell-1}$, $\Children(u_{\ell-1})(=K)$ is computed as follows:
(i) compute range $[i,j]$ on $\ArC_\ell$ and $\ArB_\ell$ as $i \gets \Select(\ArB_\ell,u)$ and $j \gets \Select(\ArB_\ell,u+1) - 1$ and
(ii) compute the set $K$ of pairs of child id $v_\ell$ and edge label $c$ as $c \gets \ArC_\ell[v]$ for all $v\in [i,j]$.

The bottom of \fref{fig:list} shows an example for computing $\Children(6_2)$.
$K = \{ (6_3,\Str{c}), (7_3,\Str{d}) \}$.
Given node $6_3$, range $[i,j]$ is computed as $i \gets \Select(\ArB_3,6) = 6$ and $j \gets \Select(\ArB_3,7) - 1 = 7$.
Pairs $(6_3,\Str{c})$ and $(7_3,\Str{d})$ in $K$ are computed as $\ArC_3[6] = \Str{c}$ and $\ArC_3[7] = \Str{d}$, respectively.

The space usage of \LIST{} for nodes at level $\ell$ is $(b+1) \cdot t_{\ell} + o(t_{\ell})$ bits.
When all $t$ nodes are represented by \LIST{}, the space usage is $\sum^{L}_{\ell=1} \{ (b+1) \cdot t_{\ell} + o(t_{\ell}) \} = (b + 1) \cdot t + (t)$ bits and is more space-efficient than the LOUDS-trie of 
$(b+2) \cdot t + o(t)$ bits of space.

\emph{Selection of \DHT{} and \LIST{} Representations:}
Either the \DHT{} representation or \LIST{} representation is adaptively applied to the series of nodes according to node density (Eq. \ref{eq:dense}) at each level.
For the selection, we ignore the auxiliary spaces for \Rank{} and \Select{} because they are negligible.
Since the space usages for the \DHT{} representation (the \LIST{} representation) at level $\ell$ are $2^b\cdot t_{\ell-1}$ ($(b+1)\cdot t_{\ell}$), we use threshold $2^b/(b+1)$ for the node density. 
If $D(\ell-1,\ell)>2^b/(b+1)$, \DHT{} is more space efficient and is applied; otherwise, \LIST{} is applied. 

\subsection{Representation for Sparse Layer}
\begin{figure}[tb]
    \centering
    \includegraphics[scale=0.35]{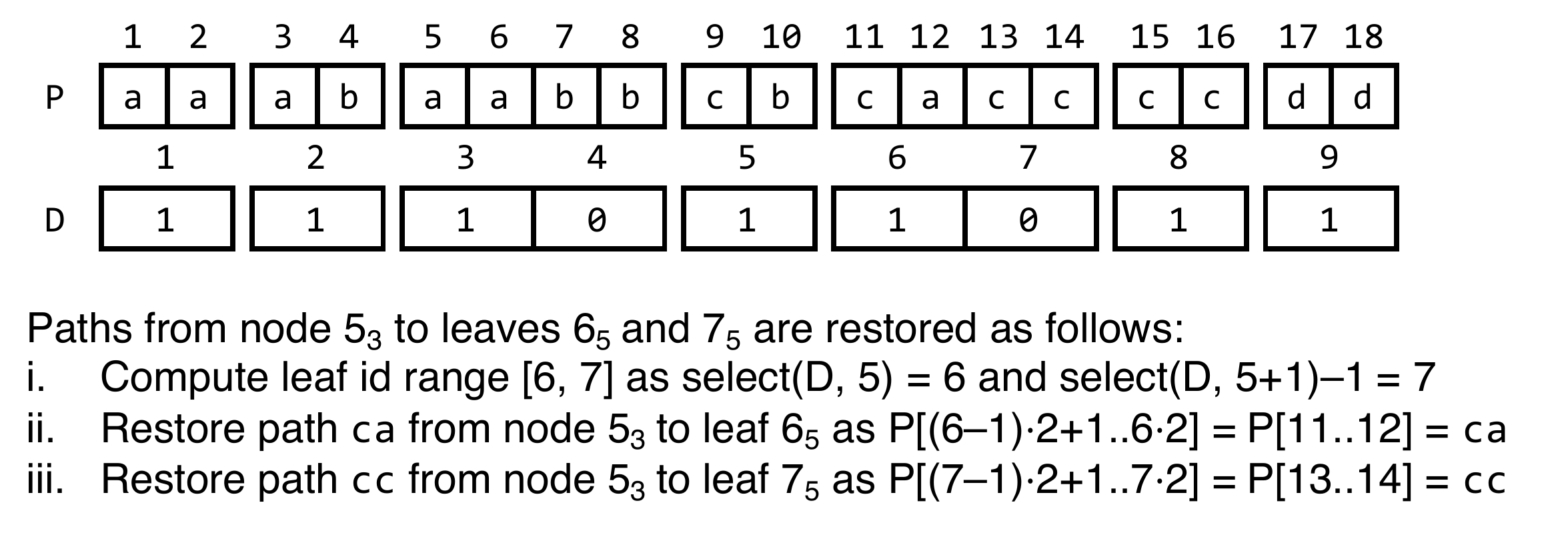}
    \caption{Representation of subtries in sparse layer of the trie in \fref{fig:bST}.}
    \label{fig:SL}
\end{figure}

The sparse layer between level $\ell_s$ and level $L$ includes a set of subtries, each of which has a height of $L - \ell_s$.
We collapse the subtries into their root-to-leaf paths and handle them as strings rather than trie structures.
The representation for the sparse layer uses two arrays of $\ArP$ and $\ArD$. 
$\ArP$ is an array of length $(L - \ell_s) \cdot t_{L}$ such that $\ArP[(L - \ell_s)\cdot (v - 1) + 1 .. (L - \ell_s) \cdot v]$ stores the edge labels on the path from the root in the subtrie containing leaf $v_L$.
$\ArD$ is a bit array of length $t_L$ such that $\ArD[v]$ stores \Str{1} if and only if leaf $v_L$ is the leftmost leaf in the subtrie.
\fref{fig:SL}, which shows arrays $\ArP$ and $\ArD$, is an example of the representation for the sparse layer. 

The paths from a given node $u_{\ell_s}$ to the leaves can be restored using the \Select{} operation on $\ArD$ in the following four steps:
(i) $i \leftarrow \Select(\ArD,u)$, 
(ii) $j \leftarrow \Select(\ArD, u+1)-1$, 
(iii) compute $\ArP[(L-\ell_s) \cdot (i-1) + 1..(L-\ell_s) \cdot j]$, and
(iv) return subarrays of $\ArP[(L-\ell_s) \cdot (i-1) + 1..(L-\ell_s) \cdot j]$ and of every length $(L-\ell_s)$ as the paths.
Leaf ids of the subtrie with root $u_{\ell_s}$ are computed as $v_{L}$ for $v \in [i,j]$. 

\fref{fig:SL} shows an example of restoring the paths \Str{ca} and \Str{cc} from node $5_3$ to leaves $6_5$ and $7_5$, respectively.
Given node $5_3$,  $i \gets \Select(\ArD, 5) = 6$ and $j \gets \Select(\ArD,6)-1 = 7$ are computed.
Leaf ids of the subtrie with root $5_3$ are $6_{5}$ and $7_{5}$.
In this example, $L - \ell_s = 5 - 3 = 2$. 
The path from root $5_3$ to leaf $6_5$ is $\ArP[(2 \cdot (6-1) + 1)..(2 \cdot 6)] = \ArP[11..12] = \Str{ca}$.
The path from root $5_3$ to leaf $7_5$ is $\ArP[(2 \cdot (7-1) + 1)..(2 \cdot 7)] = \ArP[13..14] = \Str{cc}$.

After restoring paths in a subtrie by using $\ArP$ and $\ArD$, we can simulate traversing the subtrie by computing the Hamming distance between the query and the paths.
Therefore, the fast computation of $\Ham$ is crucial in the sparse layer, 
which is explained in the next paragraph. 

\emph{Hamming Distance Computation Approach:}
Let us consider computing $\Ham(s,q)$ for $b$-bit sketches $s$ and $q$, each of length $L$.
When the operation is naively performed by comparing $s$ and $q$ character by character, the computation time is $\Order(L)$.

\begin{figure}[tb]
    \centering
    \includegraphics[scale=0.55]{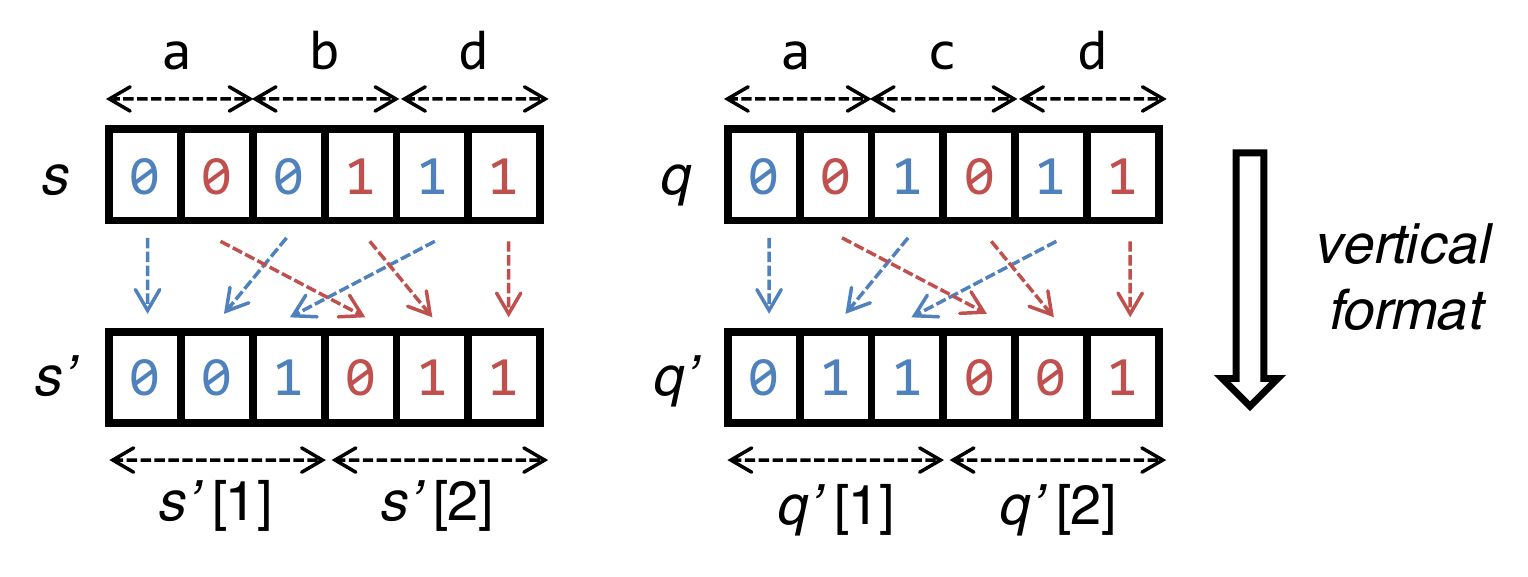}
    \caption{Illustration of the sketches $s'$ and $q'$ in vertical format for 2-bit sketches of length 3. $s = \Str{abd}$, and $q = \Str{acd}$.}
    \label{fig:fastham}
\end{figure}

Zhang et al. \cite{zhang2013hmsearch} proposed a faster computation approach by exploiting a vertical layout and the bit-parallelism offered by CPUs.
For this approach, we consider the binary representation of a character in sketches, e.g., $\Str{a} = \Str{00}$, $\Str{b} = \Str{01}$, $\Str{c} = \Str{10}$, and $\Str{d} = \Str{11}$ for $b=2$.
The approach encodes $s$ into $s'$ in a \emph{vertical} format, i.e., the $i$-th significant $L$ bits of each character of $s$ are stored in $s'[i]$ of consecutive $L$ bits.
The resulting $s'$ is an array of length $b$ in which each element is in $L$ bits.
\fref{fig:fastham} shows an example of 2-sketches each of length 3, which are represented in the vertical format.

Given sketches $s'$ and $q'$, we can compute $\Ham(s,q)$ as follows.
Initially, we prepare a bitmap $bits$ of $L$ bits in which all the bits are set to \Str{0}.
For each $i = 1,2, \ldots, b$, we iteratively perform $bits \gets bits \vee (s'[i] \oplus q'[i])$, where $\vee$ and $\oplus$ denote bitwise OR and XOR operations, respectively.
For the resulting $bits$, $\Pop(bits)$ is the same as $\Ham(s,q)$, where $\Pop$ counts the number of $\Str{1}$s and belongs to the instruction sets of any modern CPU.
For example, for $s'$ and $q'$ in \fref{fig:fastham}, the resulting $bits$ becomes $\Str{010}$ as $(s'[1] \oplus q'[1]) \vee (s'[2] \oplus q'[2]) = (\Str{001} \oplus \Str{011}) \vee (\Str{011} \oplus \Str{001}) = \Str{010} \vee \Str{010} = \Str{010}$.
$\Pop(\Str{010})$ is the same as $\Ham(\Str{abd}, \Str{acd})$ (i.e., one).
The operations $\vee$, $\oplus$, and \Pop{} can be performed in $\Order(1)$ time per machine word.
Let $w$ be the machine word size; we can compute $\Ham(s,q)$ in $\Order(b \Ceil{L/w})$ time.

We conducted preliminary experiments to compare the computation speeds of the naive and vertical-format approaches.
From the result for 32-dimensional 4-bit sketches, the computation time of the vertical-format approach was more than an order of magnitude faster than that of the naive approach.
Therefore, we apply the vertical-format approach to $\ArP$ representation.

\section{Experiments}
\label{sect:ex}

In this section, we demonstrate the effectiveness of similarity searches using $b$ST through experiments using real-world datasets.
The source code implementing our $b$ST is available at \url{https://github.com/kampersanda/integer_sketch_search}.

\subsection{Setup}

We used four real-world datasets as shown in \tref{tab:dataset}.
\emph{\Book{}} is 12,886,488 book reviews in English from Amazon \cite{mcauley2013hidden}.
We eliminated the stop words from the reviews and then represented each of reviews as a 9,253,464 dimensional fingerprint where each dimension of the fingerprint represents the presence or absence of a word.
We used $b$-bit minhash \cite{li2010b} to convert each binary vector into a 2-bit sketch of 16 dimensions.
\emph{\CP{}} consists of 216,121,626 compound-protein pairs each of which is represented as a binary vector of 3,621,623 dimensions.
We used $b$-bit minhash to convert each binary vector into a 2-bit sketch of 32 dimensions.
\emph{\SIFT{}} consists of 128 dimensional SIFT descriptors built from the BIGANN dataset \cite{jegou2011searching} of one billion images.
We used 0-bit CWS \cite{li20150} to convert each feature into a 4-bit sketch of 32 dimensions.
\emph{\GIST{}} consists of 384 dimensional GIST descriptors built from 79,302,017 tiny images \cite{torralba200880}. 
We used 0-bit CWS to convert each descriptor into a 8-bit sketch of 64 dimensions.
Following  \cite{li2010b,li2017linearized}, we used parameter settings of $b = 2$ for $b$-bit minhash and $b = 4$ or $8$ for 0-bit CWS.

\begin{table}
    \centering
    \caption{Summary of datasets.}
    \label{tab:dataset}
    \begin{tabular}{lrlrr}
        \toprule
         & $n$ & Hashing & $L$ & $b$ \\
        \midrule
        \Book{} & 12,886,488 & $b$-bit minhash & 16 & 2 \\
        \CP{} & 216,121,626 & $b$-bit minhash & 32 & 2 \\
        \SIFT{} & 1,000,000,000 & 0-bit CWS & 32 & 4 \\
        \GIST{} & 79,302,017 & 0-bit CWS & 64 & 8 \\
        \bottomrule
    \end{tabular}
\end{table}

\begin{table}
\centering
\caption{Average number of solutions.}
\label{tab:solutions}
\begin{tabular}{lrrrrr}
\toprule
 & $\tau = 1$ & $\tau = 2$ & $\tau = 3$ & $\tau = 4$ & $\tau = 5$ \\
\midrule
\Book{} & 12 & 28 & 181 & 1,273 & 7,671 \\
\CP{} & 418 & 622 & 1,473 & 2,831 & 5,201 \\
\SIFT{} & 174 & 1,057 & 5,603 & 26,840 & 111,727 \\
\GIST{} & 168 & 1,664 & 10,787 & 51,085 & 189,188 \\
\bottomrule
\end{tabular}
\end{table}

We randomly sampled 1,000 vectors from each dataset for queries.
We evaluated the search time for $\tau$ in the range from 1 to 5 and chose parameters $L$ for each dataset in consideration of the number of solutions obtained.
\tref{tab:solutions} shows the average number of solutions for each $\tau$, and a substantial number of solutions are obtained.

We conducted all experiments on one core of quad-core Intel Xeon CPU E5--2680 v2 clocked at 2.8 Ghz in a machine with 256 GB of RAM, running the 64-bit version of CentOS 6.10 based on Linux 2.6.

\subsection{Comparison of Succinct Tries}
\label{sect:ex:trie}

We compared $b$ST with the state-of-the-art succinct tries of LOUDS-trie and FST in  combination with the single-index approach.
LOUDS-trie was implemented using the TX library downloadable from \url{https://github.com/hillbig/tx-trie}, and 
FST was implemented using the SuRF library downloadable from 
\url{https://github.com/efficient/SuRF}. 
Since the single-index approach needs the inverted index for similarity searches, 
it enables us to 
fairly evaluate the search performance and space usage of data structures for implementing an inverted index.
Thus, we evaluated the performance of similarity searches on single-index using a succinct trie on each dataset.  
We fixed parameter $\lambda = 0.5$. 
Parameters $\ell_m$ and $\ell_s$ were used as 
$(\ell_m,\ell_s) = (8,11)$ for \Book{}, 
$(\ell_m,\ell_s) = (9,14)$ for \CP{},
$(\ell_m,\ell_s) = (0,21)$ for \SIFT{}, and
$(\ell_m,\ell_s) = (0,49)$ for \GIST{}.

LOUDS-trie and FST were applicable to sketches whose length was less than $2^{32}$ because of the implementation issues with the TX and SuRF libraries, respectively. 
Thus, they were not applicable to \SIFT{} whose length was 32 billion. 

\begin{table}[t]
\centering
\caption{
Average search time in milliseconds per query and space usage in mebibytes (MiB) of succinct trie.
The fastest time and smallest space are in bold.
}
\label{tab:trie}
\begin{tabular}{lrrrrrr}
\toprule
& \multicolumn{5}{c}{\Book{}} \\
& $\tau=1$ & $\tau=2$ & $\tau=3$ & $\tau=4$ & $\tau=5$ & \multicolumn{1}{c}{Space} \\
\cmidrule(lr){2-6}
\cmidrule(lr){7-7}
$b$ST & \textbf{0.006} & \textbf{0.05} & \textbf{0.37} & \textbf{2.0} & \textbf{8.3} & \textbf{9} \\
LOUDS & 0.036 & 0.32 & 2.23 & 11.6 & 45.2 & 24 \\
FST & 0.021 & 0.19 & 1.42 & 7.7 & 31.5 & 18 \\
\midrule
& \multicolumn{5}{c}{\CP{}} \\
& $\tau=1$ & $\tau=2$ & $\tau=3$ & $\tau=4$ & $\tau=5$ & \multicolumn{1}{c}{Space} \\
\cmidrule(lr){2-6}
\cmidrule(lr){7-7}
$b$ST & \textbf{0.019} & \textbf{0.14} & \textbf{0.88} & \textbf{4.9} & \textbf{23} & \textbf{308} \\
LOUDS & 0.090 & 0.67 & 4.37 & 23.1 & 99 & 741 \\
FST & 0.084 & 0.60 & 3.88 & 20.0 & 82 & 548 \\
\midrule
& \multicolumn{5}{c}{\SIFT{}} \\
& $\tau=1$ & $\tau=2$ & $\tau=3$ & $\tau=4$ & $\tau=5$ & \multicolumn{1}{c}{Space} \\
\cmidrule(lr){2-6}
\cmidrule(lr){7-7}
$b$ST & \textbf{0.22} & \textbf{3.4} & \textbf{30} & \textbf{171} & \textbf{690} & \textbf{6,082} \\
LOUDS & -- & -- & -- & -- & -- & -- \\
FST & -- & -- & -- & -- & -- & -- \\
\midrule
& \multicolumn{5}{c}{\GIST{}} \\
& $\tau=1$ & $\tau=2$ & $\tau=3$ & $\tau=4$ & $\tau=5$ & \multicolumn{1}{c}{Space} \\
\cmidrule(lr){2-6}
\cmidrule(lr){7-7}
$b$ST & \textbf{0.32} & \textbf{3.8} & \textbf{26} & \textbf{105} & \textbf{304} & \textbf{1,072} \\
LOUDS & 0.75 & 9.5 & 65 & 279 & 905 & 1,329 \\
FST & 0.53 & 7.0 & 49 & 206 & 633 & 1,163 \\
\bottomrule
\end{tabular}
\end{table}

\tref{tab:trie} shows the experimental results of search time and space usage.
$b$ST was much faster than LOUDS-trie and FST by a large margin. 
$b$ST was at most 6.2 times faster than LOUDS-trie and was at most 3.8 times faster than FST on \Book{}.
$b$ST was at most 5.0 times faster than LOUDS-trie and was at most 4.4 times faster than FST on \CP{}. 
$b$ST was much more space-efficient than LOUDS-trie and FST. 
$b$ST was 2.6 times smaller than LOUDS-trie and was 1.9 times smaller than FST on \Book{}. 
$b$ST was 2.4 times smaller than LOUDS-trie and was 1.8
times smaller than FST on \CP{}.

Those results show that $b$ST as an engineered representation for $b$-bit sketches 
was much more efficient than LOUDS-trie and FST with respect to search performance and space usage. 
$b$ST was the only method applicable to \SIFT{}. 
The similarity search performance for $b$ST is shown in the next subsection.

\subsection{Comparison of Similarity Search Methods}
\label{sect:ex:comp}
We compared single-index and multi-index using $b$ST with representative similarity search 
methods of \HashS{}, \HashM{}, and \Hm{}. 
Single-index and multi-index using $b$ST for implementing an inverted index are referred to as \TrieS{} and \TrieM, respectively. 
\Hm{} is the state-of-the-art similarity search for $b$-bit sketches and is reviewed in \sref{sect:relate}.

\HashS{} and \HashM{} are single-index and multi-index using the hash table for implementing an inverted index, respectively. 
Since \HashS{} and \HashM{} were designed for binary sketches (i.e., $b=1$) as in \cite{norouzi2014fast}, we modified
them for integer sketches (i.e., $b > 1$).
Since the similarity search of \HashS{} with large $\tau$ and $b$ took
a large amount of time, we limited the execution time to within 10 seconds per query.

The implementations of \HashS{} and \HashM{} are also contained in our library.
The implementation of \Hm{} is available at \url{https://github.com/kampersanda/hmsearch}.

For \TrieM{} and \HashM{}, we tested the number of blocks $m \in \{2,3,4\}$ for each threshold $\tau$ and chose the best value of $m$ achieving the fastest similarity search. 
\TrieM{} with $m = 2$ was the fastest for all pairs of datasets and thresholds.
\HashM{} with $m = 3$ was the fastest for $\tau \in [4,5]$ on \GIST{}, and \HashM{} with $m = 2$ was the fastest for the other pairs.

\begin{figure*}[tb]
    \centering
    \includegraphics[scale=0.5]{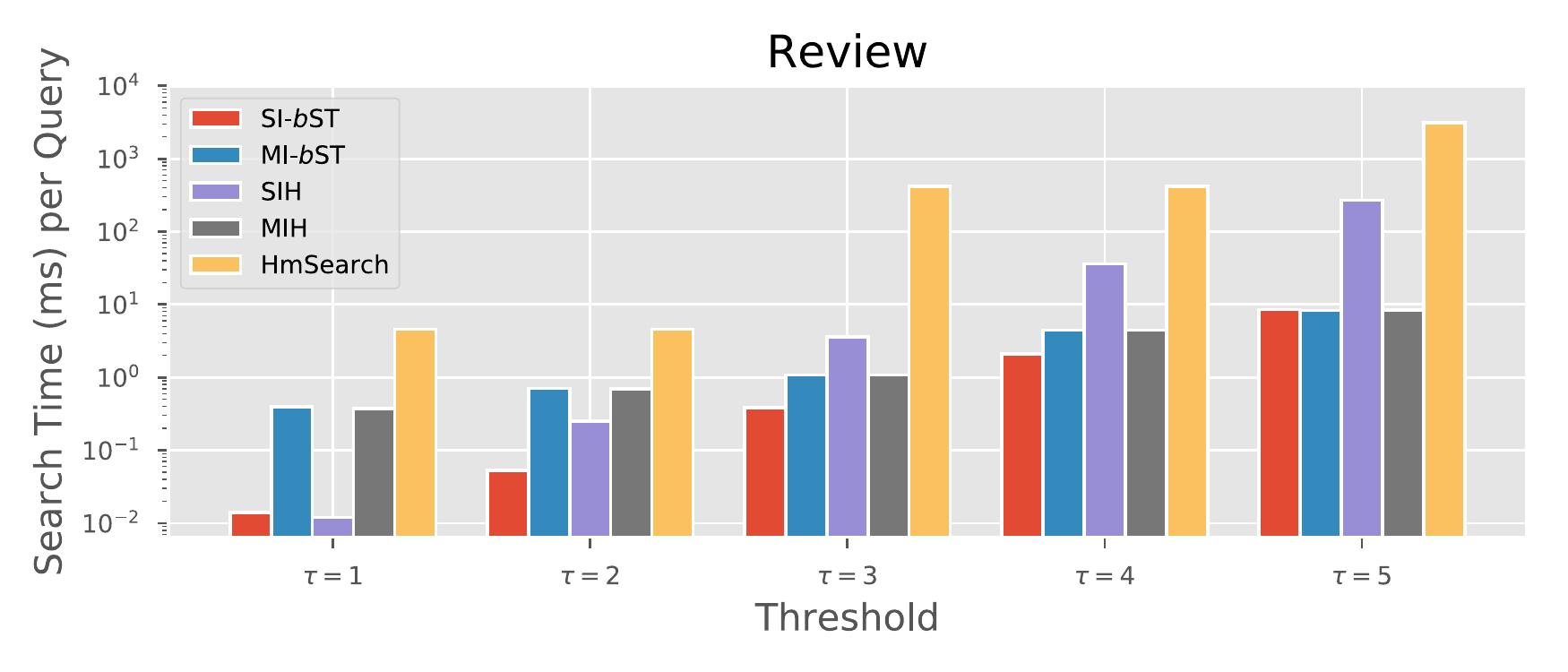}
    \includegraphics[scale=0.5]{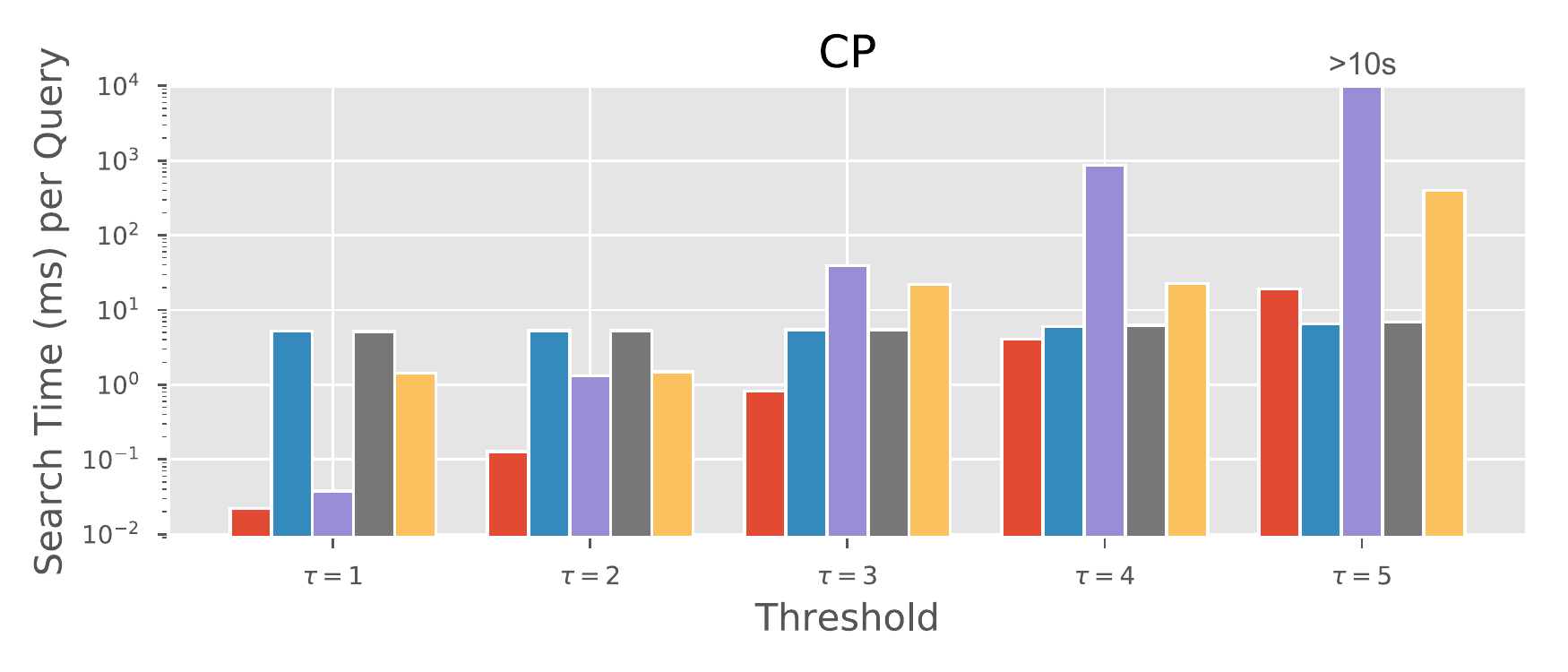}\\
    \includegraphics[scale=0.5]{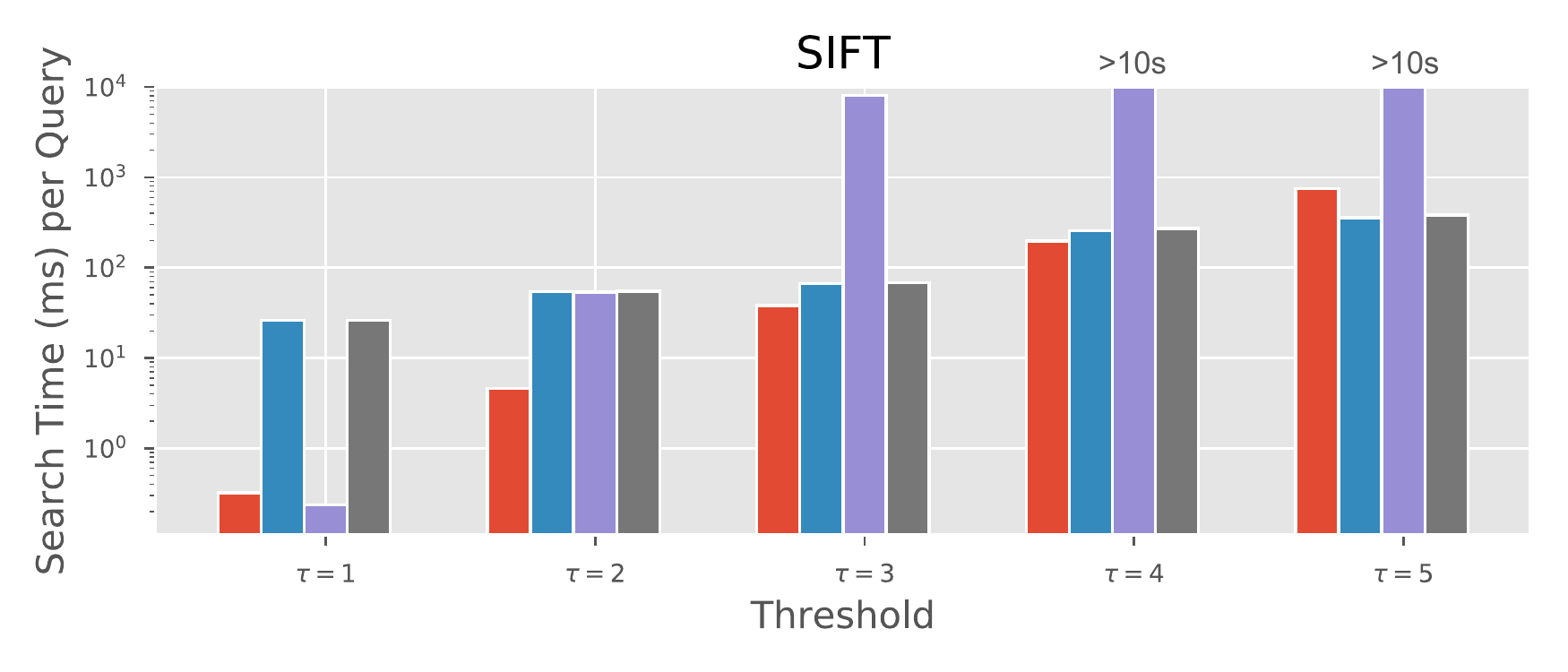}
    \includegraphics[scale=0.5]{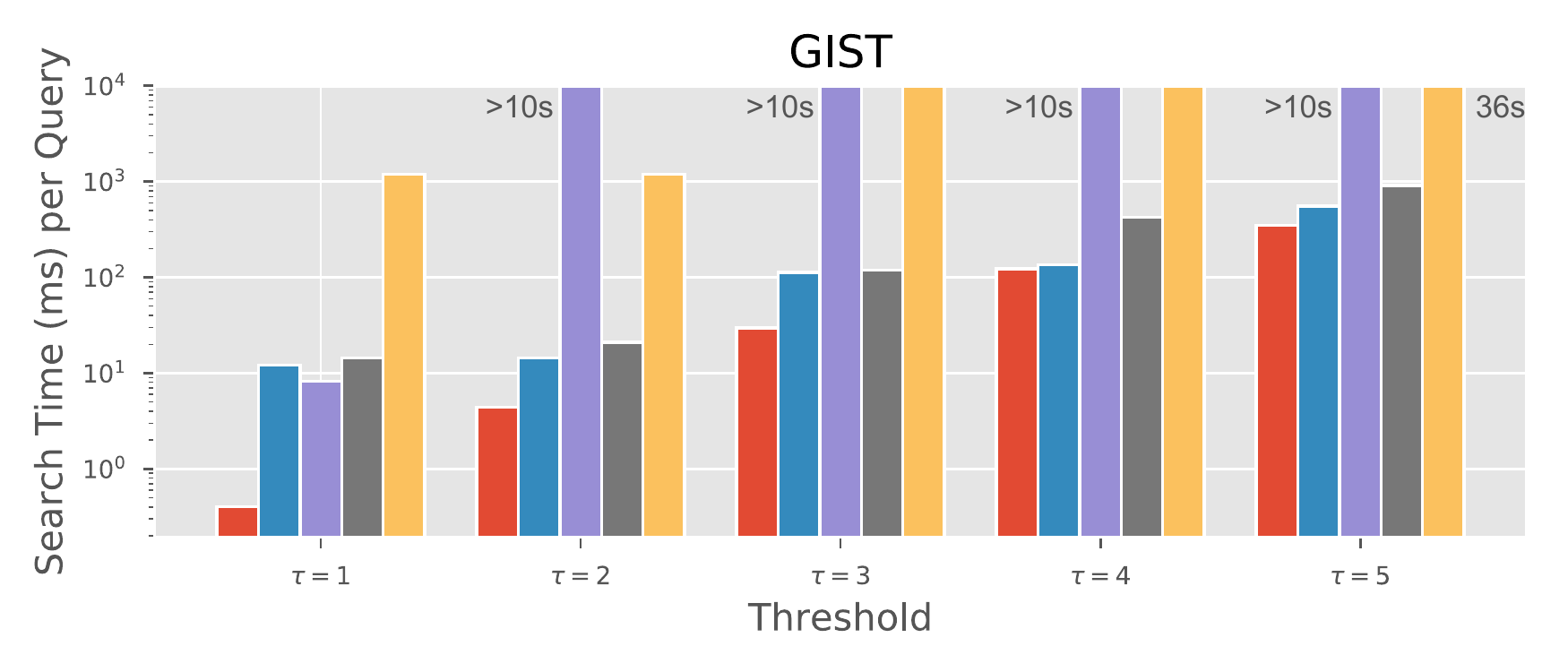}
    \caption{Average search time in milliseconds per query (log scale).
    The bars denote \TrieS{} (red), \TrieM{} (blue), \HashS{} (purple), \HashM{} (gray), and \Hm{} (yellow) starting from the left.
    The upper limit of results the figure plots is 10 seconds since we aborted the similarity search process of \HashS{} if the time exceeds 10 seconds. Other than \HashS{}, only the result of \Hm{} for \GIST{} when $\tau = 5$ exceeded 10 seconds (it was 36 seconds).}
    \label{fig:results:query}
\end{figure*}

\fref{fig:results:query} shows the average similarity search time in milliseconds (ms) per query for each method. 
Since \HashS{} had been designed for binary sketches, its performance was evaluated only with them~\cite{torralba2008small,norouzi2014fast}. 
In contrast, when integer sketches were used, \HashS{} did not perform well even for small $\tau$ for each 
dataset. \HashS{} did not finish within $10$ seconds for a $\tau$ of no less than four on \SIFT{} and no less than two on \GIST{} because the large numbers of signatures were generated on those datasets. 

Although \HashM{} had also been designed for binary sketches as well as \HashS{}, 
it performed well for large $\tau$ in contrast to \HashS{}. 
On the other hand, \HashM{} was slower than \HashS{} for small $\tau$ (e.g., $\tau=1$) on each dataset. 

Although \Hm{} was a similarity search on the multi-index approach with engineered assignments of the number of blocks and 
thresholds, it was slower than \HashM{} on \Book{}, \CP{}, and  \GIST{}. 
In addition, the space usage of \Hm{} exceeded the memory limitation of 256 GB on \SIFT{}. Those results on $b$-bit sketches were consistent with those results on binary sketches, which 
were shown in \cite{qingeneralizing}.

For each $\tau \leq 4$, \TrieS{} was the fastest among all the methods on each dataset, while only \HashS{} was competitive compared to \TrieS{} for $\tau = 1$ on \Book{} and \SIFT{}. 
For $\tau = 5$, \TrieM{} and \HashM{} were the fastest and competitive except on \GIST{}, while \TrieS{} was the fastest on \GIST{}.
The results show our \TrieS{} and \TrieM{} were the fastest for similarity searches 
on huge datasets of $b$-bit sketches. 



\tref{tab:space} shows the experimental results of space-efficiency.
\Hm{}, \HashM{}, and \TrieM{} were similarity searches on the multi-index approach.
\Hm{} consumed a large amount of memory and consumed approximately 860 MiB memory for \Book{}, more than 256 GiB memory for \SIFT{}, and at least 25 GiB memory for \GIST{}. 
\HashM{} was more space-efficient than \Hm{}, but \HashM{}'s space-usage was problematic for 
large datasets.
In particular, \HashM{} consumed more than 26 GiB of memory for \SIFT{} and more than 5.7 GiB of memory 
for \GIST{}.
\TrieM{} was the most space-efficient among all methods 
on the multi-index approach for each dataset.
\TrieM{} consumed 3.2 GiB for \CP{}, 23 GiB for \SIFT{}, and 5.4 GiB for \GIST{}.
Although \HashS{} was a similarity search on the single-index approach, \HashS{} was 
not space-efficient, and it consumed 2.3 GiB for \CP{} , 32 GiB for \SIFT{}, and 4.5 GiB for \GIST{}. \TrieS{} was the most space-efficient among all the methods for each dataset. 
\TrieS{} consumed only 1.0 GiB for \CP{}, 9.6 GiB for \SIFT{}, and 1.3 GiB for \GIST{}. 

For thresholds $\tau \leq 4$, the results of the comparison of similarity searches showed that \TrieS{} was the best among all the methods in terms of search time and space usage.
For $\tau = 5$, \TrieM{} can be used instead of \TrieS{} for fast and space-efficient similarity searches.

\begin{table}[tb]
\centering
\caption{
Space usage of similarity search method in MiB.
The smallest space usage is in bold.
}
\label{tab:space}
\begin{tabular}{lrrrr}
\toprule
& \Book{} & \CP{} & \SIFT{} & \GIST{} \\
\midrule
\TrieS{} & \textbf{48} & \textbf{1,057} & \textbf{9,802} & \textbf{1,338} \\
\TrieM{} $(m=2)$ & 126 & 3,232 & 23,159 & 5,513 \\
\HashS{} & 172 & 2,329 & 32,727 & 4,501 \\
\HashM{} $(m=2)$ & 125 & 4,633 & 28,876 & 6,128 \\
\HashM{} $(m=3)$ & 160 & 3,997 & 26,665 & 5,744 \\
\Hm{} ($\tau=1,2$) & 866 & 53,097 & -- & 48,456 \\
\Hm{} ($\tau=3,4$) & 860 & 29,396 & -- & 27,337 \\
\Hm{} ($\tau=5$) & 860 & 28,866 & -- & 25,305 \\
\bottomrule
\end{tabular}
\end{table}

\section{Conclusion}
We presented $b$ST, a novel succinct representation of trie for fast and space-efficient similarity searches on $b$-bit sketches. 
Our experimental results using real-world datasets demonstrated that $b$ST outperformed other state-of-the-art succinct tries in terms of search time and memory. 

Subsequently, we presented \TrieS{} and \TrieM{}, single-index and multi-index using $b$ST implementing an inverted index. 
Our experimental results demonstrated that \TrieS{} was the fastest for thresholds $\tau \leq 4$. 
For $\tau=5$, \TrieM{} was the alternative to \TrieS{}. 
In addition, the space-efficiency of \TrieS{} was the best among all the methods, and 
it consumed 10 GiB of memory for storing a billion-scale database, while a state-of-the-art method consumed 29 GiB of memory.

\bibliographystyle{IEEEtran}
\bibliography{bibfiles/library_exorg}

\appendix
\subsection{Evaluation for Single- and Multi-indexes}
\label{app:sih}

We evaluate the time performance of the single-index approach using a hash table data structure (i.e., SIH) for the similarity search per query as cost $cost_S$ in the following equation: 
\begin{equation}
\label{eq:cost_s}
    cost_S = \Sigs(b,L,\tau) \cdot L + |I|, 
\end{equation}
where $\Sigs(b,L,\tau)$ is the number of signatures as follows:
\begin{eqnarray}
\label{eq:sig1}
    \Sigs(b,L,\tau) = \sum^{\tau}_{k=0} \binom{L}{k}(2^b - 1)^{k}.
\end{eqnarray}
Since $cost_S$ depends largely on the number of signatures, it 
is linearly proportional to $L$ and exponentially proportional to $\tau$ and $b$.


We evaluate the time performance of the multi-index approach using hash tables (e.g., MIH) for the similarity search per query as cost $cost_M$ by 
the summation of the filtering and verification costs as follows: 
\begin{equation}
\label{eq:cost-v}
    cost_M = \sum^m_{j=1} \left\{ \Sigs(b,L^j,\tau^j) \cdot L^j + L \cdot |C^j| \right\}, 
\end{equation}
where term $\sum^m_{j=1} \Sigs(b,L^j,\tau^j) \cdot L^j$ is the filtering cost for $m$ blocks of query, and term $L\sum^m_{j=1} |C^j|$ is the verification cost depending on the total number of candidate solutions and sketch length. 

\begin{figure}[tb]
\centering
\includegraphics[scale=0.42]{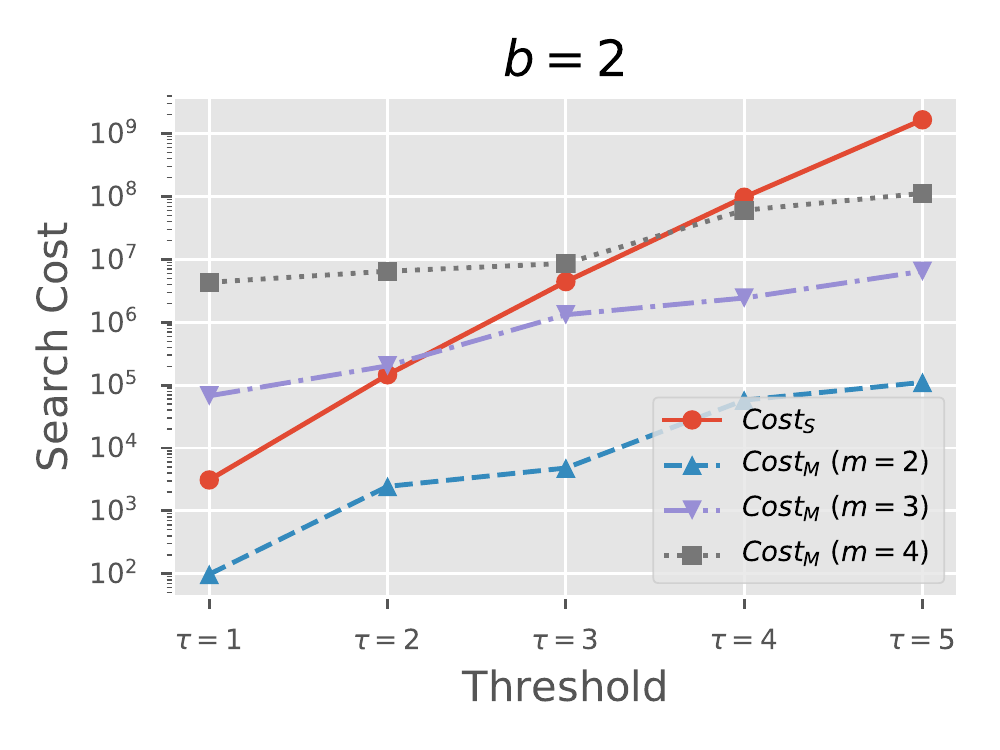}
\includegraphics[scale=0.42]{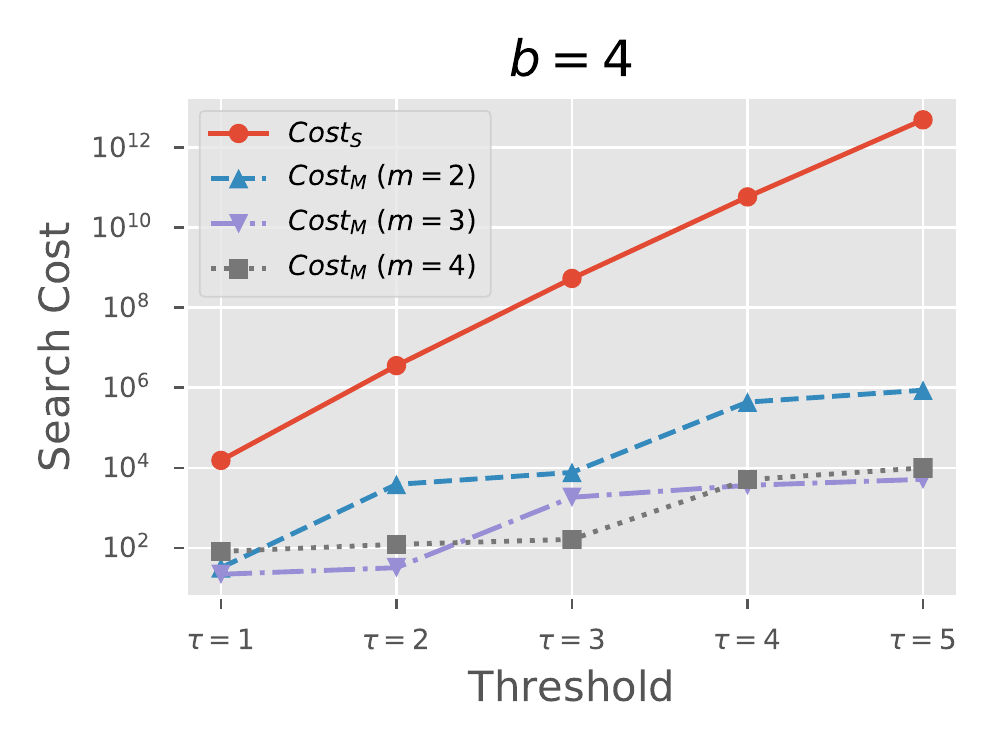}
\caption{$cost_S$ and $cost_M$ for $b=2$~(left) and $b=4$~(right); the other parameters are chosen as  $L = 32$, $b \in \{2,4\}$, and $m \in \{2,3,4\}$.}
\label{fig:cost}
\end{figure}

\fref{fig:cost} shows the values of costs $cost_S$ and $cost_M$ by fixing the parameters of  $(n,L)=(2^{32},32)$ and varying the parameters of $m\in \{2,3,4\}$ and $\tau\in \{1,2, \ldots ,5\}$. 
In that figure, we compute $|I| = \Sigs(b,L,\tau) \cdot n / (2^b)^{L}$ in $cost_S$ and $|C^j| = \Sigs(b,L^j,\tau^j) \cdot n / (2^b)^{L^j}$ in $cost_M$ 
under the assumption that $n$ sketches in a database are uniformly distributed 
in the Hamming space.

We can see $cost_S$ for the single-index approach exponentially increases for 
parameters $\tau$ and $b$. Thus, similarity searches on the single-index approach 
cannot be applied to $b$-bit sketches and large $\tau$.
We can also see that $cost_M$ for the multi-index approach increases for parameters of $\tau$ and $b$, but 
the increase is relatively small when large $m$ (e.g., $m=4$) is used. 
However, when a large number of blocks are used, a large number of candidate solutions are generated, 
resulting in a large verification cost.
In addition, since many inverted indexes are built for large blocks, methods on the multi-index approach consume a large amount of memory. 

\end{document}